\documentclass[11pt]{article}

\usepackage[a4paper, left=0.7in, right=0.7in, top=1.1in, bottom=1.1in]{geometry}
\usepackage[utf8]{inputenc}
\usepackage[T1]{fontenc}
\usepackage{amsmath, amssymb, amsthm}
\usepackage{graphicx}
\usepackage{url}
\usepackage{hyperref}
\usepackage{authblk}  

\usepackage{mathtools}
\mathtoolsset{showonlyrefs,showmanualtags}
\usepackage{physics}
\usepackage{multirow}
\usepackage{xcolor}
\usepackage{listings}
\usepackage{booktabs}
\usepackage{makecell}

\title{CALT: A Library for Computer Algebra with Transformer}

\author[1,2]{Hiroshi Kera\thanks{Email: \href{mailto:kera@chiba-u.jp}{kera@chiba-u.jp}}}
\author[1]{Shun Arakawa}
\author[1]{Yuta Sato}

\affil[1]{Chiba University}
\affil[2]{Zuse Institute Berlin (ZIB)}

\date{}  %

\def\cD{{\mathcal{D}}}

\def\cT{{\mathcal{T}}}

\def\bF{{\mathbb{F}}}

\def\bZ{{\mathbb{Z}}}

\theoremstyle{plain} \numberwithin{equation}{section}
\newtheorem{theorem}{Theorem}[section]
\numberwithin{theorem}{section}

\theoremstyle{definition}

\newtheorem{example}[theorem]{Example}

\theoremstyle{plain}

\newcommand{\ring}{k[x_1,\ldots, x_n]}

\newcommand{\ideal}[1]{\langle{#1}\rangle}

\newcommand{\fs}{f_1,\ldots, f_s}
\newcommand{\gs}{g_1,\ldots, g_t}
\newcommand{\xs}{x_1,\ldots, x_n}

\usepackage[ruled, linesnumbered]{algorithm2e} %
\usepackage{xcolor} % 
\usepackage{tikz}
\usetikzlibrary{fit,calc}
\colorlet{pink}{red!40}
\colorlet{lightblue}{blue!30}
\colorlet{lightgreen}{green!30}

\DontPrintSemicolon

\begin{document}

\maketitle

\begin{abstract}
Recent advances in artificial intelligence have demonstrated the learnability of symbolic computation through end-to-end deep learning. Given a sufficient number of examples of symbolic expressions before and after the target computation, Transformer models---highly effective learners of sequence-to-sequence functions---can be trained to emulate the computation. This development opens up several intriguing challenges and new research directions, which require active contributions from the symbolic computation community. In this work, we introduce \emph{Computer Algebra with Transformer} (CALT), a user-friendly Python library designed to help non-experts in deep learning train models for symbolic computation tasks.
\end{abstract}

\section{Introduction}
Recent success in deep learning suggests new directions in symbolic computation. Transformer models~\cite{vaswani2017attention} are deep learning architectures that efficiently learn sequence-to-sequence functions. Originally developed for natural language processing (i.e., texts as sequences of words), they have since been applied to various domains, including computer vision (i.e., images as sequences of pixels or patches)~\cite{vit2021an} and game agents (i.e., sequences of states and actions)~\cite{chen2021decision}. Notably, several studies have explored the use of Transformer models to learn symbolic computations. Successful examples include symbolic integration~\cite{lample2020deep}, mid-scale code breaking~\cite{wenger2022salsa}, Lyapunov function design~\cite{alfarano2024global}, and Gr\"obner basis computation~\cite{kera2024learning}.

Deep learning of symbolic computations not only offers a new approach to computationally expensive problems but also raises novel challenges in computer algebra. For example, training a Transformer model to compute Gr\"obner bases requires a large training set $\cD = \{(F_i, G_i)\}_i$, consisting of pairs where $F_i \subset \ring$ is a generating set and $G_i \subset \ring$ is a Gr\"obner basis of the ideal $\ideal{F_i}$. The efficient generation of such samples poses new problems. In the forward approach—where $\{F_i\}_i$ are first sampled and then $\{G_i\}_i$ are computed—how can we randomly generate diverse $\{F_i\}_i$ that yield non-trivial ideals? In the backward approach, how can we sample diverse Gr\"obner bases $\{G_i\}_i$ and then transform them into generating sets $\{F_i\}_i$ such that $\ideal{F_i} = \ideal{G_i}$ for each $i$\,?

A recent study~\cite{kera2024learning} presented an algorithm for generating such examples when the ideals are in shape position, which was further examined from a geometric perspective in~\cite{kambe2025geometric}. Beyond the dataset generation problem, other promising directions include integrating Transformer models into computer-algebraic algorithms, or analyzing the computational difficulty of symbolic problems via the learning difficulty. All these problems clearly invite contributions from the symbolic computation community, but many researchers in this field are unfamiliar with training Transformer models. 

In this paper, we introduce a new library, \textbf{C}omputer \textbf{Al}gebra with \textbf{T}ransformer (\textbf{CALT}), which provides an accessible framework for learning symbolic computations using Transformer models. The library is designed to be user-friendly for computer algebraists and non-experts in deep learning. In the minimal use case, users are only required to implement a small function that randomly generates a computing instance, such as an input--output pair $(F, G)$. A built-in data generation helper can then execute this function in parallel to efficiently construct a dataset $\cD = \{(F_i, G_i)\}_i$. 

\textbf{CALT} offers the following key features:\footnote{The code is available at \url{https://github.com/HiroshiKERA/calt}}.
\begin{itemize}
    \item It is implemented in Python and SageMath~\cite{sagemath}. The Python backend enables seamless integration with deep learning libraries such as PyTorch~\cite{pytorch} and HuggingFace Transformers~\cite{transformers}, while SageMath provides extensive support for symbolic operations and compatibility with existing computer algebra systems. 
    \item Users can alternatively supply training instances as plain text files, using a domain-specific language for sample generation, since the training pipeline also supports text-based input formats.
    \item For advanced use, \textbf{CALT} provides modular components that allow users to customize symbolic preprocessing, model architectures, and training workflows.
\end{itemize}

\section{Learning symbolic computation.}
We here introduce the end-to-end learning of symbolic computation using Transformer models.  
We denote by $\Sigma$ the set of basic symbols that constitute the sequences (e.g., letters, numerals, and operators), and $\Sigma^{*}$ denotes the set of all finite strings over $\Sigma$.

\subsection{Transformer Model}\label{sec:transformer-model}

We briefly introduce a Transformer model and its training procedure. We provide only an overview and omit technical details.

The Transformer can be regarded as a sequence-to-sequence function parameterized by $\theta \in \mathbb{R}^D$:
\begin{align}\label{eq:transformer-simple}
  \mathcal{T}_{\theta} \colon \Sigma^{*} \to \Sigma^{*}.
\end{align} 
Training a Transformer reduces to optimizing the parameters $\theta$ to minimize the loss over a training set $\mathcal{D} = \{(X_i, Y_i)\}_{i=1}^m$, where $X_i, Y_i \in \Sigma^*$:
\begin{align}
    \min_{\theta \in \mathbb{R}^D} \mathcal{L}(\mathcal{T}_{\theta},\mathcal{D}), \ \ \text{with}\ \ \mathcal{L}(\mathcal{T}_{\theta},\mathcal{D}) = \frac{1}{m} \sum_{i=1}^m \ell(\mathcal{T}_{\theta}(X_i), Y_i),
\end{align}
where the loss function $\ell: \Sigma^* \times \Sigma^* \to \mathbb{R}_{\ge 0}$ measures the proximity between the prediction $\mathcal{T}_{\theta}(X_i)$ and the ground truth $Y_i$.
This optimization is performed via gradient-based iterative updates of $\theta$:
\begin{align}
    \theta \leftarrow \theta - \lambda \, \frac{\partial}{\partial \theta}\mathcal{L}(\mathcal{T}_{\theta}, \mathcal{D}),
\end{align}
where $\lambda > 0$ is the step size. The \emph{backpropagation} mechanism~\cite{Goodfellow-et-al-2016} enables efficient evaluation of the gradient at a given $\theta$.

\subsection{Tokenization}\label{sec:tokenization}
The Transformer is trained with a pair of input and output sequences, $(X, Y) \in \Sigma^* \times \Sigma^*$. We present the conversion process of symbolic functions to the sequences. In this context, the set $\Sigma$ is called the \textit{vocabulary}, and its entries are called \textit{tokens}. We denote by $\Sigma^*$ the set of all sequences of tokens. The process of obtaining a sequence of tokens $X \in \Sigma^*$ is called \textit{tokenization}.

\begin{example}[Multiplication of polynomials]
Let $f_1, f_2 \in \mathbb{Z}[x_1,x_2]$ be two polynomials whose product
$f_1 \cdot f_2$ constitutes the learning target. The vocabulary may be defined by 
\begin{align*}
    \Sigma
    = \{\text{E0},\ldots,~\text{Es},\;
       \text{C0},~\ldots,~\text{Ct},\;
       +,\;-,\;\mid,\;>,\;<\}.
\end{align*}
Here, E$a$ and C$b$ represent,
respectively, an exponent value $a$ and a coefficient value $b$.
For instance, the monomial $x_1^{2}x_2^{3}$ corresponds to
$(\text{C1},\text{E2},\text{E3})$.
The symbols $\mid,\,>,\,<$ are widely used \textit{special tokens} denoting
a separator, the beginning of a sequence, and the end of a sequence,
respectively.  For $(f_1,f_2)=(x_1+x_2,\,x_1-x_2)$,
we have sequences of tokens 
\begin{align*}
    \pi(f_1)
      &= (>, \text{C1},\text{E1},\text{E0},
          +,\text{C1},\text{E0},\text{E1}, <),\\
    \pi(f_1,f_2)
      &= (>, \text{C1},\text{E1},\text{E0},
          +,\text{C1},\text{E0},\text{E1},
          \mid,\text{C1},\text{E1},\text{E0},
          -,\text{C1},\text{E0},\text{E1}, <).
\end{align*}
The map $\pi$ is called a \textit{tokenizer}. 
\end{example}

In practice, users may implement a custom vocabulary and tokenizer as needed. For example, when working over the polynomial ring $\mathbb{R}[\xs]$, a coefficient, say $c = 1.23$, can be tokenized in multiple ways, such as  
$(\text{C1.23})$ or $(\text{C1}, ., \text{C2}, \text{C3})$.  
The former requires a large vocabulary to cover many distinct coefficient tokens, while the latter only needs ten digit tokens (plus the decimal point).  

The Transformer $\mathcal{T}_{\theta}$ maintains an \textit{embedding vector} for each token, which is updated during training and eventually encodes relationships among tokens.  
A large vocabulary involves many embedding vectors, increasing the total number of parameters (i.e., the dimension $D$ of $\theta \in \mathbb{R}^D$).  
Meanwhile, it is known that the Transformer's memory cost grows quadratically with the input sequence length.  
Therefore, digit-level coefficient tokenization such as $(\text{C1}, \text{C2}, \text{C3})$ incurs higher memory usage during training and inference. The tokenization and the design of embedding vectors impact the computational cost and learning efficiency; \cite{charton2022linear} tested several methods of tokenizing real-value coefficients, and \cite{golkar2023xval} proposed to introduce number token \text{[C]} and its embedding vector $v_{\text{[C]}}$, and define the embedding vector of C$b$ as its scaling, i.e., $bv_{\text{[C]}}$.

\subsection{Next-token prediction}
In Section~\ref{sec:transformer-model}, we introduced the Transformer as a mapping from a sequence to sequence, Eq.~\eqref{eq:transformer-simple}. To be more precise, it receives two sequences as input. 
\begin{align}
    \cT_{\theta}: \Sigma^* \times \Sigma^* \to \Sigma^*.
\end{align}
Let $X = (>, x_1, \ldots, x_s, <) \in \Sigma^s$ and $Y = (>, y_1, \ldots, y_t, <)\in \Sigma^t$ be tokenized sequences. 
The Transformer receives $X$ and generates $Y$.
Importantly, this generation is realized by the \textit{next-token prediction}; for $k=0, \ldots t$, the Transformer predicts $y_{k}$ from $X$ \textit{and} its previous predictions as folows. 
\begin{align}
    \hat{y}_1 &= \cT_{\theta} (X, Y_0),\quad \text{with}\quad Y_0 = (>), \\
    \hat{y}_2 &= \cT_{\theta} (X, Y_1),\quad \text{with}\quad Y_1 = (>, \hat{y}_1), \\
    \hat{y}_3 &= \cT_{\theta} (X, Y_2),\quad \text{with}\quad Y_2 = (>, \hat{y}_1, \hat{y}_2).
\end{align}
This continues until the end-of-sequence token $<$ is predicted. This process is called \textit{auto-regressive generation} as well as \textit{chain-of-thought} (CoT). Such a step-by-step prediction is not only flexible to the output length, but also gives a strong reasoning ability~\cite{shen2023positional,kim2025transformers}.
We will observe this in Section~\ref{sec:showcase}.

\section{Learning Pipeline}
\begin{figure}[t]
  \centering
  \includegraphics[width=0.6\linewidth]{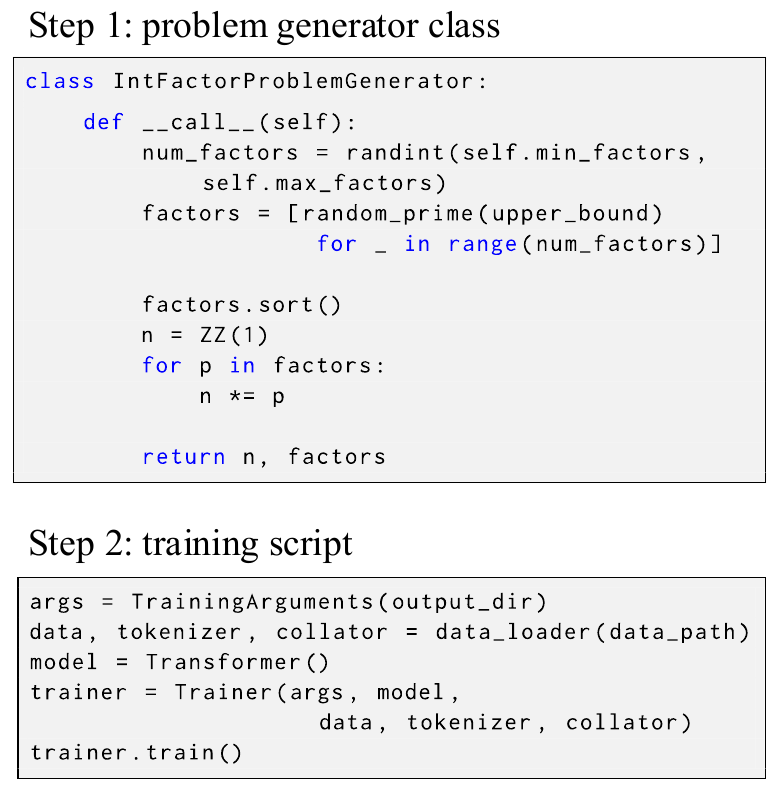}
  \caption{Problem generation and training pipeline. The dataset is created by a user-defined class and subsequently used by the training script.}
  \label{fig:generate_train_scripts}
\end{figure}

To train a Transformer for a symbolic task, the following components must be implemented:

\begin{enumerate}
    \item \textbf{Instance generator.}  
    This function generates an input--output pair $(F, G)$ corresponding to the target symbolic computation.  
    Efficiently generating such pairs $(F, G)$ \emph{without} explicitly performing the target computation often raises interesting and underexplored problems in computer algebra.
    
    \item \textbf{Tokenizer.}  
    Once a symbolic instance $(F, G)$ is obtained, it must be converted into a pair of token sequences $(\pi(F), \pi(G))$.  
    To do this, one must define both the vocabulary $\Sigma$ and a tokenizer $\pi \colon \Sigma^* \to \text{Tokens}^*$.

    \item \textbf{Transformer trainer.}  
    Given the tokenized input--output pairs, one can train a Transformer model by preparing its architecture and executing the training loop, while logging relevant metrics and outputs.
\end{enumerate}

Our library, \textbf{CALT}, enables users to focus primarily on the mathematical and practical design of (1) \textbf{instance generation functions}, while also offering customizable modules for (2) \textbf{tokenization} and (3) \textbf{model training}.  
\textbf{CALT} is built on top of PyTorch~\cite{pytorch} and HuggingFace~\cite{transformers}, and is therefore functionally a subset of those frameworks.  
However, the high degree of flexibility provided by these libraries can be overwhelming for non-experts.  
CALT addresses this by providing a streamlined and purpose-specific interface tailored for learning symbolic computations. Recently, a simple codebase was proposed mainly for arithmetic tasks~\cite{charton2025int2int}.

\section{Showcases}\label{sec:showcase}

\begin{table*}[t]
\centering
\caption{Input--output samples from each task.}
\scriptsize
\begin{tabular}{p{2.0cm}|l|l}
\toprule
\textbf{Task} & \textbf{Input} & \textbf{Output}\\
\midrule
\makecell[l]{Integer \\ factorization} &
$f = 108606433$ &
$g_1 = 13,\; g_2 = 37,\;g_3 = 43,\;g_4 = 59,\;g_5 = 89$\\
\midrule
\multirow{3}{*}{\makecell[l]{Products in \\ $\mathbb{Z}[x,y]$}} &
  $f_1 = -x - 2y - 1$ & \multirow{3}{*}{$\displaystyle
  g = f_1f_2f_3 = 2x^4y + 4x^3y^2 - 2x^4 - 2x^3y - x^2y^2 - 2xy^3 - 2x^3 + x^2y + xy^2 + xy$} \\
& $f_2 = -2x^2 + y$ & \\
& $f_3 = xy - x$ & \\
\midrule
\multirow{3}{*}{\makecell[l]{Products in \\ $\mathbb{F}_7[x,y]$}} &
  $f_1 = -2x^2 - 3x + 2$ & \multirow{3}{*}{$\displaystyle
  g = f_1f_2f_3 = -2x^5 - 2x^4y + 3x^3y^2 + x^4 - x^3y + x^2y^2 - x^3 - 3x^2y - 3xy^2 - 3x^2 + x + y + 3$} \\
& $f_2 = x^2 - 3xy + 1$ & \\
& $f_3 = x - 3y - 2$ & \\
\midrule
\multirow{3}{*}{\makecell[l]{Products in \\ $\mathbb{F}_7[x,y]$ \\ with CoT}} &
  $f_1 = -2x^2 - 3x + 2$ & $g_1 = f_1 = -2x^2 - 3x + 2$\\
& $f_2 = x^2 - 3xy + 1$ & $g_2 = f_1f_2 = -2x^4 - x^3y - 3x^3 + 2x^2y + xy - 3x + 2$\\
& $f_3 = x - 3y - 2$ & $g_3 = f_1f_2f_3 = -2x^5 - 2x^4y + 3x^3y^2 + x^4 - x^3y + x^2y^2 - x^3 - 3x^2y - 3xy^2 - 3x^2 + x + y + 3$\\
\bottomrule
\end{tabular}
\label{tab:dataset_samples}
\end{table*}

We now provide several examples of learning symbolic computations using our library on the following arithmetic and symbolic tasks. 
We denote the input and output of the target symbolic computation respectively by tuples $F = (\fs)$ and $G=(\gs)$, the entries of which are either integers or polynomials. Polynomials are supposed to be in fully expanded form.

\begin{itemize}
    \item \textbf{Prime factorization:} Transformer receives an integer $f \in \bZ$ and predict its prime factorization $f = g_1\cdots g_t$ with primes $\gs \in \bZ$ in the ascending order.
    \item \textbf{Polynomial multiplication:} Transformer receives polynomials $\fs \in k[x, y]$ and predict their product $g = f_1\cdots f_s$. We consider $k \in \{\bZ, \bF_7\}$. For $k = \bF_7$, we additionally consider the task of cumulative product, where Transformer predicts $g_1 := f_1, ~ g_2 := f_1f_2, ~ \ldots, ~ g_s := f_1\cdots f_s$.
\end{itemize}

\paragraph{User implementations.} 
Figure~\ref{fig:generate_train_scripts} shows example codes for instance generations that users should implement for each task. 
The instance generating functions are then simply taken over to the generation helper to create samples in parallel. 
Here, the user defines a custom \texttt{ProblemGenerator} class whose constructor fixes the task specification; for polynomial problems it additionally accepts a sampler that draws random polynomials from the chosen ring.
Each call to \texttt{call()} returns an input–output pair 
$(F,G)$, and iterating this routine produces the entire dataset.
After the dataset has been generated, model training is enabled by adjusting the tokenization setting (e.g., number of variables, maximum coefficients, and degree) to match the symbols present in the data.
We generate training and test sets for four tasks: i) prime factorization, ii) polynomial multiplication over $\bZ$, iii) polynomial multiplication over $\bF_7$, (iv) polynomial multiplication over $\bF_7$ with CoT.

\paragraph{Dataset generation.}
\textbf{(Prime factorization)}. 
We sample $t$ distinct primes $\{g_1,\dots,g_t\}$ from primes up to~$100$, where $t$ is uniformly chosen from $\{2,\dots,5\}$.  
We then set the input to $f=\prod_{i=1}^{t}g_i$ and use the sorted tuple $(g_1,\dots,g_t)$ as the target.  
\textbf{(Polynomial multiplication)}. 
Each factor \(f_i\) is a bivariate polynomial of total degree \(\le2\) containing at most three monomials. In task~(ii), coefficients are drawn uniformly from \(\{-2,\dots,2\}\) , whereas in Tasks~(iii) and~(iv), all coefficients are sampled from $\bF_7$.
The model receives the multiset $\{f_1,\dots,f_s\}$ and must output the fully expanded product $g=\left(\prod_{i=1}^{s}f_i\right)$.
The generation procedure in Tasks~(iii) and~(iv) is identical to Task~(ii) except that every coefficient is reduced modulo~\(7\).  
Working in $\bF_7$ removes issues with coefficient explosion while preserving the combinatorial structure of the task.
In Task (iv) we keep the same factor distribution as in Task (iii), but the target is now the ordered sequence of partial products $G=(g_i)_{i=1}^s, g_i = \prod_{j=1}^i f_j$.
This chain-of-thought (CoT) formulation encourages the model to predict intermediate results—and thus to expose its reasoning—rather than producing only the final product.

\paragraph{Training setup.} For each task, a Transformer was trained on 100,000 samples and tested on 1,000 samples. We used the standard architecture of Transformer~\cite{vaswani2017attention} with AdamW optimizer with $(\beta_1, \beta_2) = (0.9, 0.999)$ 
The learning rate was initially set at $5.0\times10^{-5}$ and then linearly decayed during the 80k  training steps. 
The batch size was set to 128, and the dropout rate was set to 0.1.

\begin{figure}[t]
  \centering
\includegraphics[width=0.5\linewidth]{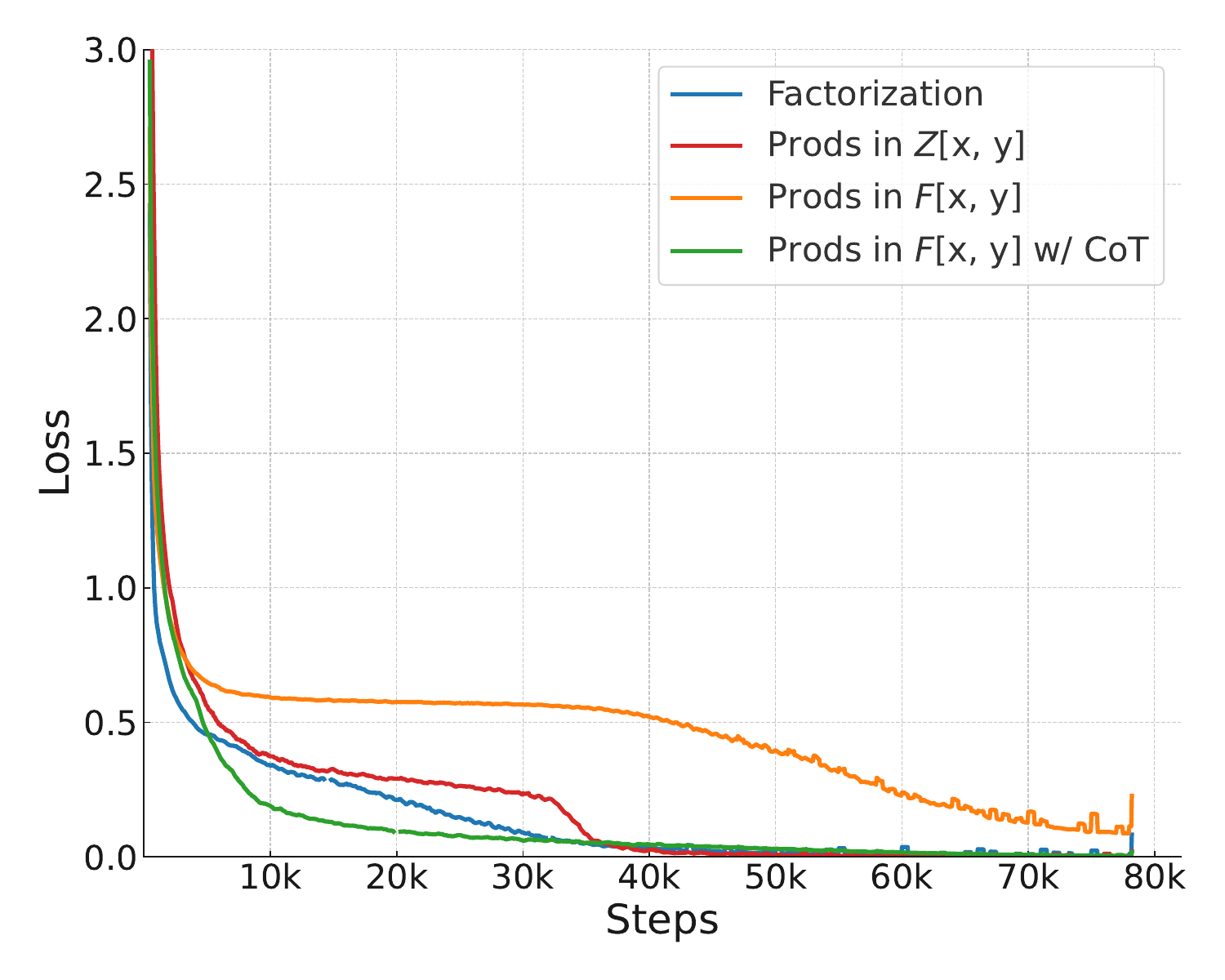}
  \caption{Loss curves for each task during training.}
  \label{fig:task_loss_curves}
\end{figure}

\paragraph{Results.}
Table~\ref{tab:exp_acc} reports the success rates of Transformers in generating complete solutions.
Even on the particularly challenging examples listed in Table~\ref{tab:dataset_samples}, the model attains a success rate exceeding 70\%.
We also observe a marked drop in performance when the task is performed over the $\bF_7$ as opposed to the $\bZ$ case.
This degradation is consistent with earlier findings that Transformers struggle to learn computations involving modular arithmetic \cite{saxena2024teachingtransformers}.
Nevertheless, when we equip the model with CoT, the accuracy over $\bF_7$ improves substantially, mitigating much of the gap.
This is because CoT allows the model to decompose the original task into a sequence of simpler subproblems, solving them step by step.  
Such a breakdown reduces the overall complexity of the task and guides the model through intermediate reasoning steps, thereby facilitating more accurate predictions.
Figure~\ref{fig:task_loss_curves} visualizes the training-loss trajectories for all four tasks. 
Tasks~(i) and~(iv) exhibit steady convergence, whereas Tasks~(ii) and~(iii) sometimes display a stagnation phase followed by a sharp decrease in loss.
This phenomenon is frequently observed in polynomial-related tasks, indicating the necessity of allocating ample optimization steps.

\begin{table}[t]
\centering
\caption{Success rate for each task.}
\begin{tabular}{l|c}
\toprule
\textbf{Task} & \textbf{Success rate} \\
\midrule
Integer factorization & $74.5\,\%$ \\
Products in $\bZ[x,y]$ & $81.2\,\%$ \\
Products in $\bF_7[x,y]$ & $31.7\,\%$ \\
Products in $\bF_7[x,y]$ w/ CoT & $74.2\,\%$ \\
\bottomrule
\end{tabular}
\label{tab:exp_acc}
\end{table}

\section{Upcoming updates} \textbf{CALT} is still preliminary, and several updates are planned.  
We aim to improve both dataset generation and Transformer training to make them even more user-friendly.  
Customizable tokenization pipelines will be provided for broad use cases. 
We also plan to include monomial-based embedding~\cite{kera_pelleriti2025computational}, which represents polynomials as significantly shorter yet expressive sequences.  
For users without access to GPUs, we will also provide Google Colab notebooks for basic usage.

\paragraph{Acknowledgments.}
The authors are grateful to Kohei Ohsato~(LY Corporation) and Kosyuke Sumiyasu (ZOZO, Inc.) for their valuable suggestions in designing \textbf{CALT}.
This research was partially supported by JST PRESTO Grant Number JPMJPR24K4 and the Chiba University IAAR Research Support Program and the Program for Forming Japan's Peak Research Universities (J-PEAKS).

\bibliographystyle{acm}

\end{document}